\documentclass[twocolumn]{fairmeta}

\usepackage[most]{tcolorbox}
\usepackage{amsmath}
\usepackage{amssymb}
\usepackage{amsfonts}
\usepackage{mathtools}
\usepackage{nicefrac}
\usepackage{bbding}
\usepackage{colortbl}
\usepackage{wrapfig}
\usepackage{float}
\usepackage{soul}

\definecolor{color1}{RGB}{255,250,205}
\definecolor{color2}{RGB}{255,228,225}
\definecolor{color3}{RGB}{34,139,34}
\definecolor{hku_color}{HTML}{13A983}
\definecolor{szu_color}{HTML}{84193E}

\usepackage{xspace}

\title{RoboMirror: Understand Before You Imitate for Video to Humanoid Locomotion}

\author[1,2,\dagger]{Zhe Li}
\author[3,4,*]{Boan Zhu}
\author[4,*]{Yangyang Wei}
\author[5]{Shuanghao Bai}
\author[6]{Yuheng Ji}
\author[4]{Yibo Peng}
\author[7]{Tao Huang}
\author[4]{Pengwei Wang}
\author[4]{Zhongyuan Wang}
\author[3]{S.-H. Gary Chan}
\author[8]{Chang Xu}
\author[4]{Cheng Chi}
\author[1,2,\ddagger]{Jianfei Yang}
\author[4,9,\ddagger]{Shanghang Zhang}

\affiliation[1]{MARS Lab}
\affiliation[2]{Nanyang Technological University}
\affiliation[3]{The Hong Kong University of Science and Technology}
\affiliation[4]{Beijing Academy of Artificial Intelligence}
\affiliation[5]{Xi'an Jiaotong University}
\affiliation[6]{Chinese Academy of Sciences}
\affiliation[7]{Shanghai Jiao Tong University}
\affiliation[8]{University of Sydney}
\affiliation[9]{Peking University}

\contribution[*]{Equal contribution}
\contribution[\dagger]{Project lead}
\contribution[\ddagger]{Corresponding author}

\abstract{
Humans learn locomotion through visual observation, interpreting visual content first before imitating actions. However, state-of-the-art humanoid locomotion systems rely on either curated motion capture trajectories or sparse text commands, leaving a critical gap between visual understanding and control. Text-to-motion methods suffer from semantic sparsity and staged pipeline errors, while video-based approaches only perform mechanical pose mimicry without genuine visual understanding.
We propose RoboMirror, the first retargeting-free video-to-locomotion framework embodying ``understand before you imitate''. Leveraging VLMs, it distills raw egocentric/third-person videos into visual motion intents, which directly condition a diffusion-based policy to generate physically plausible, semantically aligned locomotion without explicit pose reconstruction or retargeting.
Extensive experiments validate RoboMirror's effectiveness: it enables telepresence via egocentric videos, drastically reduces third-person control latency by 80\%, and achieves a 3.7\% higher task success rate than baselines. By reframing humanoid control around video understanding, we bridge the visual understanding and action gap.
}

\metadata[Project Page]{\url{https://gentlefress.github.io/RoboMirror-proj/}}

\begin{document}
\maketitle

\begin{figure*}[t]
    \centering
    \includegraphics[width=0.8\textwidth]{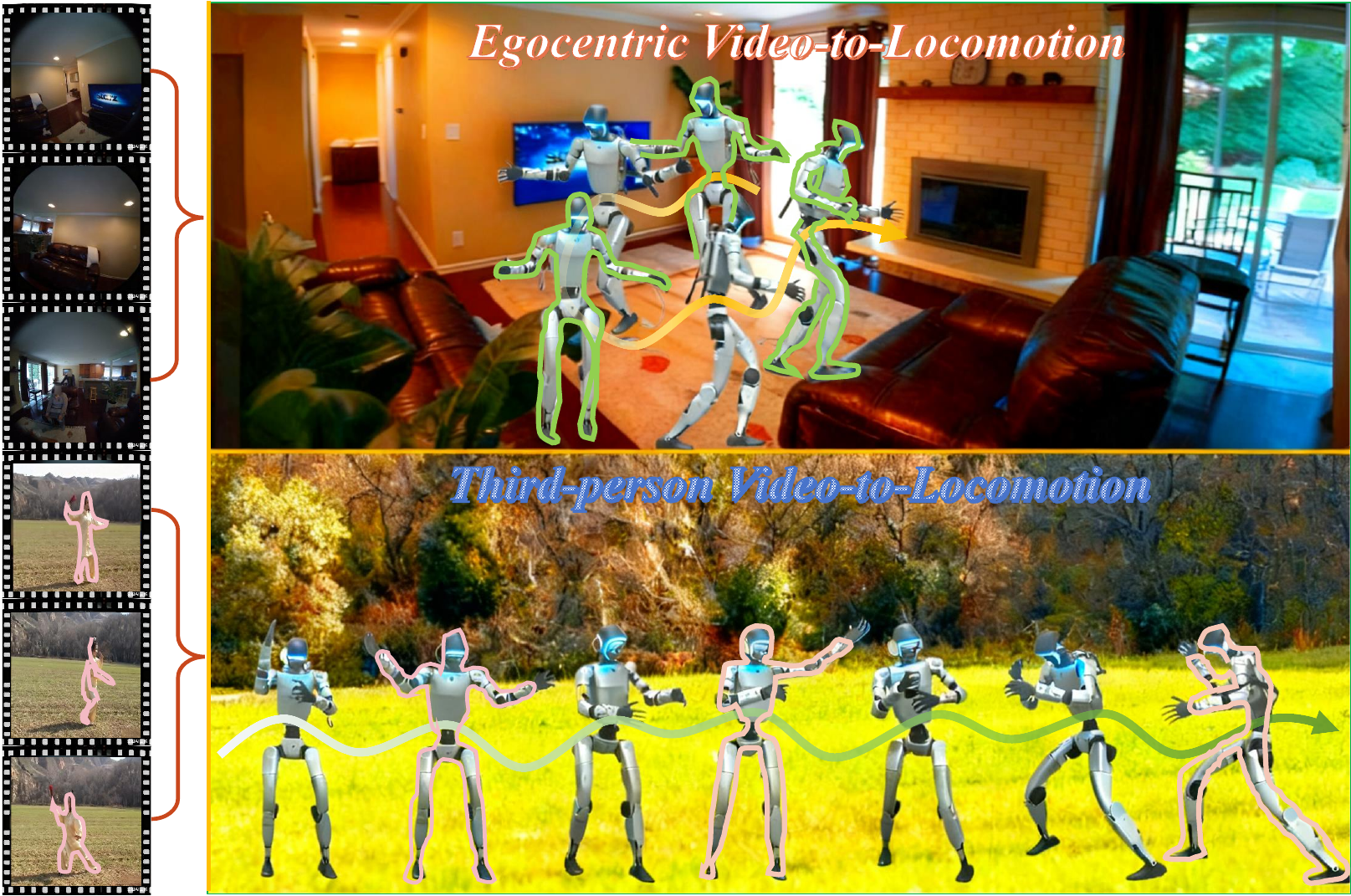}
    \caption{\textbf{RoboMirror} makes humanoids understand before imitating. It acts like a mirror: from egocentric videos it infers and replicates the camera wearer's actions from changes in the surrounding viewpoint; from third-person videos it first understands the action and then imitates it, without pose estimation or retargeting at inference time.}
    \label{fig:motivation}
\end{figure*}

\section{Introduction}
\label{sec:intro}

Humans intuitively learn locomotion by observing, a process predicated on understanding visual percepts before imitating them. 
Yet, current humanoid paradigms fail to replicate this ``understand-then-act'' architecture. 
Prevailing video-based methods degenerate into kinematic mimicry, a brittle process focused on reconstructing poses rather than inferring intent. 
Other dominant modalities are even less grounded: Motion Capture (MoCap) bypasses visual perception entirely, while text commands compress rich visual context into sparse symbols. 
A critical gap thus remains: current approaches either fail to genuinely understand rich visual data or bypass it altogether, fundamentally decoupling perception from control.

This failure to understand is exemplified by the ``kinematic mimicry'' paradigm~\cite{weng2025hdmi, he2025asap, he2024omnih2o, xie2025kungfubot}. 
Its ``pose estimate-retarget-track" pipeline is not only brittle, plagued by error accumulation and latency, but architecturally incapable of semantic understanding. 
By forcing the controller to track low-level kinematics, it precludes the learning of high-level visual intent. 
The alternative paradigms are flawed at the modality level. 
MoCap-driven systems~\cite{shao2025langwbc, serifi2024robot, yue2025rl} are inherently non-perceptive, while text-to-motion generation~\cite{guo2024momask, li2024lamp, li2024mulsmo, li2025omnimotion} suffers from the inherent sparsity of language, which cannot encode the rich dynamics and goals abundant in video.


In this work, we argue for a different interface: \textit{\textbf{video-to-locomotion}}. 
Unlike sparse modalities like text or reference poses, video constitutes a substantially more information-dense medium. 
As presented in Figure~\ref{fig:motivation}, both egocentric and third-person videos encapsulate rich environmental cues, including scene attributes, temporal dynamics, and action goals. 
\textbf{\textit{Our key insight is that this rich visual evidence must be internalized to condition a policy directly, rather than being reduced to brittle kinematics via pose estimation. }}
We therefore regard humanoid locomotion as a generative problem: given this internalized visual context, synthesize physically plausible and semantically grounded motion. 
This direct ``understand-then-act'' mapping unlocks two capabilities that prior approaches cannot provide: (1) telepresence, where first-person video demonstration guides the robot to perform the corresponding locomotion, creating an ``as if I were there'' experience; and (2) robust, retargeting-free third-person imitation, avoiding the error accumulation that plagues conventional pipelines.

We therefore propose \textbf{\textit{RoboMirror}}, a retargeting-free framework named to emphasize the mirror-like mapping from video to action.
Our core technical insight is to use a Vision-Language Model (VLM) to distill raw video into a visual latent representation, which is then explicitly trained to reconstruct a corresponding motion latent. 
This reconstructed motion latent serves as the sole conditioning signal for a diffusion-based locomotion policy. 
This architecture explicitly bridges semantic visual understanding with physics-based control. 
Concretely, we leverage the robust generalization of VLMs~\cite{Qwen2.5-VL, Qwen2-VL, Qwen-VL} to obtain visual latents from first- or third-person videos. These latents are mapped to motion latents, which are then fed into the diffusion policy to generate smooth, executable actions, bypassing explicit pose estimation and retargeting entirely.


Extensive experiments validate the effectiveness and practicality of RoboMirror. Compared directly to imitation based on pose estimation, our method dramatically accelerates the pipeline from video understanding to on-robot deployment, reducing latency from 9.22 $s$ to 1.84 $s$. Beyond sheer speed, our method delivers higher-quality control by avoiding retargeting failures, as evidenced by a 3.7\% absolute increase in task success rate and lower tracking error relative to baseline. Furthermore, for egocentric videos~\cite{ma2024nymeria}, we demonstrate that robust, semantically grounded locomotion can be synthesized without any explicit human pose supervision, a task where traditional pose-estimation pipelines fail. Crucially, we successfully introduce VLM into the humanoid control loop, enabling video-conditioned policies that can be further extended to fine-grained hand manipulation and low-friction teleoperation. In short, RoboMirror reframes humanoid control around video understanding. By learning to understand first and imitate second, we close the gap between what is seen and how to move, moving from fragile pose mimicry to robust, visually grounded action.

Our contributions can be summarized as follows:
\begin{itemize}
    \item We propose RoboMirror, the first retargeting-free framework for humanoid locomotion that replaces brittle pose reconstruction with a direct mapping from 2D video to visual motion intent, enabling a true end-to-end, video-to-locomotion policy.

    \item We introduce a VLM-assisted locomotion policy where a VLM distills visual latents from raw video. These latents are trained to reconstruct motion latents, which in turn serve as a robust, non-kinematic conditioner for a diffusion-based action generator.

    \item We validate RoboMirror's significant outperformance against ``pose estimation-retarget-track'' baselines in task success rate and latency. Crucially, we demonstrate robust locomotion from egocentric video without explicit pose supervision, a task where traditional pipelines fail.
\end{itemize}

\section{Related Work}

\subsection{Humanoid Whole-body Control}
Model-based whole-body controllers deliver precise task execution via accurate dynamics and contact modeling, but entail heavy modeling effort and limited generalization to new skills or unmodeled dynamics~\cite{geyer2003positive, sreenath2011compliant}. Learning-based approaches ease modeling burden yet hinge on carefully crafted, task-specific rewards; despite successes in challenging settings such as locomotion on complex terrains, jumping, and fall recovery~\cite{wang2025beamdojo, peng2021amp, li2023robust, huang2025learning, he2025learning, yuan2026roboforge, li2023enhancing, zhou2025roborefer, zhou2025robotracer, liu2026sapave, tan2026robobrain}, they require per-task reward engineering and often struggle to produce coordinated, human-like behaviors. To disentangle distinct upper- vs. lower-body objectives, some studies split control into independent policies~\cite{zhang2025falcon, li2025hold}, which can undermine inter-limb coordination and limit generality. Others adopt hierarchical planning to learn complex sequential skills, e.g.table tennis~\cite{su2025hitter}, improving modularity but adding design complexity and latency. Whole-body motion tracking reframes the objective by directly using human motion as supervision~\cite{han2025kungfubot2}, eliminating task-specific rewards while promoting globally coordinated, expressive motion across diverse skills. This perspective provides a unified control objective and a scalable path toward human-like whole-body behavior.

\subsection{Humanoid Motion Tracking}
Learning realistic behaviors from human motion has become central to high-fidelity control. DeepMimic~\cite{peng2018deepmimic} introduced a phase-based tracking framework with random state initialization and early termination to stabilize imitation.
To mitigate sim-to-real gaps for dynamic skills, ASAP~\cite{he2025asap} proposed a multi-stage pipeline with a delta-action model. HuB~\cite{zhang2025hub} and KungfuBot~\cite{xie2025kungfubot} further leverage sophisticated motion preprocessing and tracking mechanisms to achieve precise imitation of highly dynamic single motions.

For multi-skill policies within a single controller, OmniH2O~\cite{he2024omnih2o} demonstrated a universal policy that catalyzed subsequent research on broad motion libraries. ExBody2~\cite{ji2024exbody2} improves expressiveness via decomposed tracking targets and motion filtering. TWIST~\cite{ze2025twist} and CLONE~\cite{li2025clone} attain high-quality tracking in teleoperation settings but primarily cover lower-dynamic motions. BumbleBee~\cite{wang2025experts} employs a two-stage strategy, which is clustering motions to train expert policies, then distilling them into a unified policy. GMT~\cite{chen2025gmt} achieves robust tracking of aggressive motions by prioritizing root velocity and pose information over global positions. UniTracker~\cite{yin2025unitracker} supports dynamic tracking but relies on global targets, limiting stability on long sequences. BeyondMimic~\cite{liao2025beyondmimic} attains high-fidelity single-motion tracking via carefully designed objectives and precise system identification, followed by distillation into a unified diffusion policy for task-specific control. KungfuBot2~\cite{han2025kungfubot2} proposes an orthogonal MoE to realize a general motion tracking policy spanning diverse skills. Building on these insights, we pursue a universal policy that conditions on video to generate humanoid locomotion, moving from fragile kinematic mimicry to visually grounded performance control and transforming humanoids into semantically aligned imitators.

\subsection{Modality-driven Humanoid Locomotion}
Recent work explores conditioning humanoid locomotion on high-level modalities, with language as a prominent interface. LangWBC~\cite{shao2025langwbc} pairs a compact auxiliary network with the control policy to generate motions online from instructions, but its limited capacity hinders scaling to complex, diverse motion distributions and offers weak guarantees for generalization to unseen prompts. RLPF~\cite{yue2025rl} fine-tunes an LLM and introduces physical feasibility feedback from a motion-tracking policy to iteratively align semantic intent with executable motion, helping bridge sim-to-real; however, heavy decoder updates risk catastrophic forgetting of pretrained knowledge. RoboGhost~\cite{li2025language} proposes a latent-driven, retargeting-free framework that treats locomotion as generation, reducing error accumulation and inference latency, but it considers language as the sole input modality. RoboPerform~\cite{li2026you} proposes a retargeting-free audio-to-locomotion framework that unifies music-driven dance and speech-driven co-speech gesture generation for humanoids. LeVERB~\cite{xue2025leverbhumanoidwholebodycontrol} explores vision-language-guided whole-body control by leveraging visual feedback and linguistic instructions; it depends on retargeting pre-collected motion data and focuses mainly on quasi-static or low-dynamic tasks. In contrast, we directly process raw egocentric/third-person video without motion retargeting, aligning semantic understanding from vision with physically feasible actions using a latent-driven diffusion policy.
\section{Method}

\begin{figure*}[h]
\centering
  \includegraphics[width=1.0\textwidth]{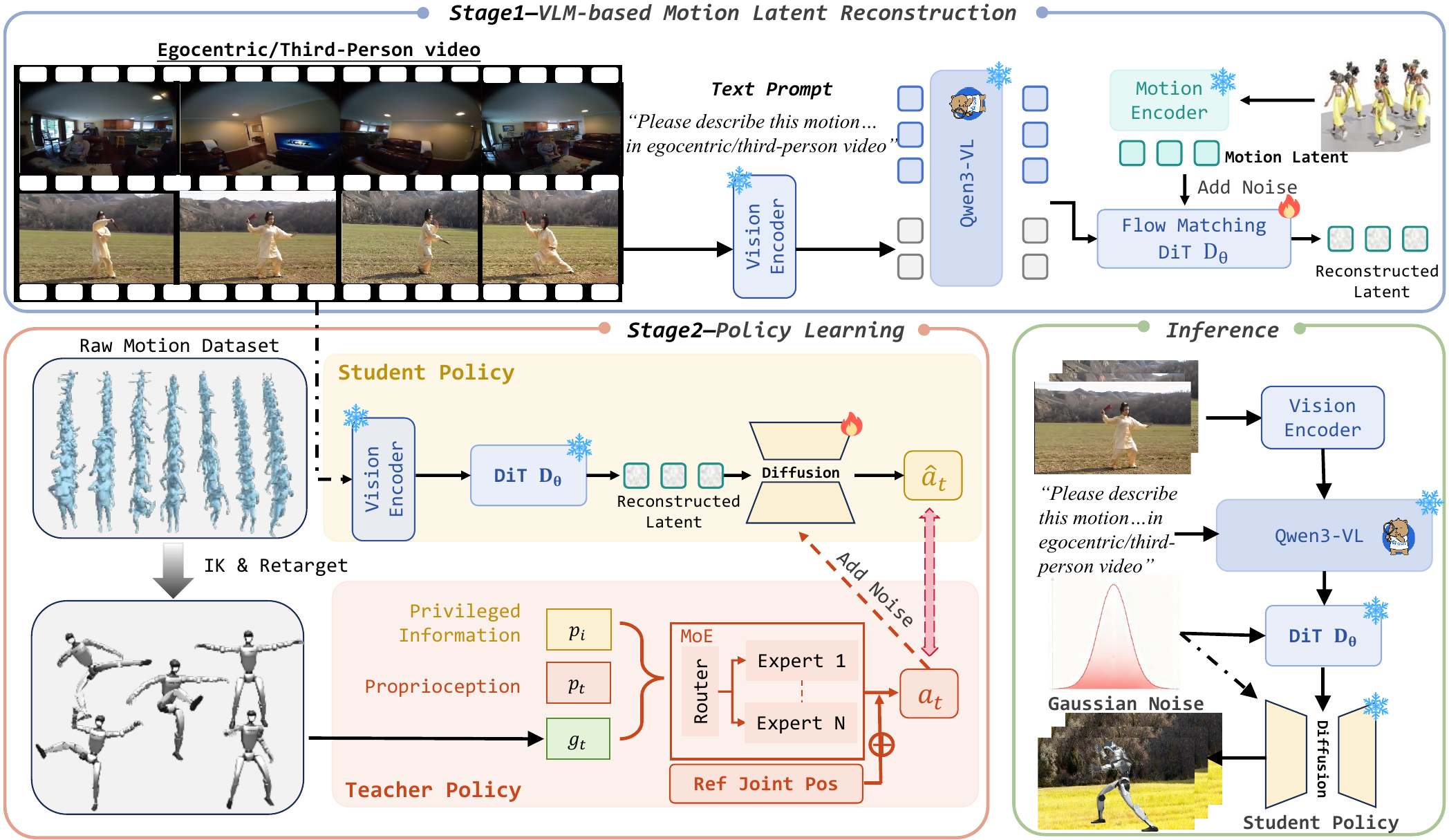}
\caption{Overview of RoboMirror. It adopts a two-stage framework: initially, it leverages Qwen3-VL to process egocentric or third-person video inputs, generating motion latents through diffusion models with DiT $\mathcal D_\theta$. Subsequently, in the policy learning stage, a MoE-based teacher policy is trained with RL, while a diffusion-based student policy learns to denoise actions under the guidance of reconstructed motion latents. During inference, it can first understand and then imitate the motion in the video without motion obtainment and retargeting.}
\label{fig:framework}
\end{figure*}

This section presents the core components of our framework, which is depicted in Figure~\ref{fig:framework}. We start with an overview of our main framework and its motivation in Section 3.1, offering a high-level description of the architectural design and underlying rationale. Section 3.2 elaborates on the method for reconstructing motion latent from VLM-derived latent, which leverages a diffusion model for accurate and robust reconstruction. Furthermore, Section 3.3 introduces our MoE-based residual teacher policy and the latent-guided diffusion-based student policy, along with a detailed exposition of their inference procedures. Other implementation details are provided in the Appendix.

\subsection{Overview}

We present a novel framework that enables robots to understand visual content from videos and execute corresponding physical actions in a coherent manner. At its core, our framework replaces simplistic cross-modal alignment with motion latent reconstruction, an intentional choice rooted in the principle that robust, semantically meaningful vision-motion mapping arises from generative reconstruction rather than superficial feature matching. As illustrated in Figure~\ref{fig:framework}, our approach comprises three sequentially linked components: a VLM-based multi-view video understanding module, a VLM-conditioned diffusion model for motion latent reconstruction, and a diffusion-based policy training module. We address a critical challenge: generating physically executable actions for humanoid robots directly from videos of varying perspectives (first- or third-person), entirely eliminating reliance on error-prone pose estimation.

The pipeline unfolds in a natural "understand-reconstruct-control" progression. It begins by feeding input videos into a pretrained VLM, which extracts a high-level semantic latent $l_{\text{VLM}}$ encoding both action details and contextual scene information. Unlike existing methods that naively repurpose $l_{\text{VLM}}$ as a control signal, we treat this latent as a semantic anchor for a diffusion-based motion latent reconstructor. This design stems from a key insight: "reconstruction outperforms alignment." When a model learns to reconstruct kinematically coherent motion latents from VLM semantics, it achieves not only more robust cross-modal alignment but also inherently embeds physical plausibility constraints, avoiding the semantic disconnection plaguing direct alignment strategies. 

The reconstructed motion latent $l_{\text{motion}}$ then fuses with the robot's proprioceptive states and real-time observations to condition a diffusion-based deployment policy $\pi$. Through iterative denoising, this policy outputs actions directly executable on humanoid platforms. By obviating explicit motion estimation and alignment procedures, our reconstruction-driven paradigm enables an end-to-end video-to-action mapping, uniquely suited for practical deployment scenarios where reliability and efficiency are paramount.

\subsection{Motion Latent Reconstruction from Vision-Language Model}
\label{vlm}
Leveraging the strong image and video understanding capabilities as well as robust generalization of vision-language models, we adopt Qwen3-VL~\cite{Qwen2.5-VL} as our video understanding module. We first train a VAE~\cite{kingma2013auto} on our motion dataset to reconstruct motion sequences. For video processing, we feed first-person or third-person videos into Qwen3-VL with task-specific prompts. For egocentric videos, the prompt is \textit{"Please describe the motion of the first-person individual in the egocentric video"} while for third-person videos it is \textit{"Please describe the motion in the video"}. This design ensures the latent representations output by Qwen3-VL are enriched with high-quality semantic information about the motion content.


For infusing kinematic information into video latents and reconstructing motion representations, we employ a flow-matching based diffusion model, denoted as $\mathcal{D}_\theta$, which takes VLM-derived video latents as conditional signals to reconstruct VAE-learned motion latents. As illustrated in Fig. \ref{fig:framework}, $\mathcal{D}_\theta$ consists of stacked transformer blocks with adaptive layer normalization, enabling effective conditioning on video semantics while preserving motion kinematic structure~\cite{bjorck2025gr00t}. The model operates on noised motion latents $\epsilon_{\text{motion}}$, with cross-attention blocks attending to the video latents $\mathbf{l}_{\text{VLM}}$ to enforce semantic consistency, and self-attention blocks capturing temporal dependencies within motion sequences.

Given a ground-truth motion latent $l_{\text{motion}}$ from pretrained VAE, a flow-matching timestep $\tau \in [0, 1]$, and sampled Gaussian noise $\epsilon \sim \mathcal{N}(0, \mathbf{I})$, the noised motion latent $\epsilon_{\text{motion}}$ is constructed as:
\[
\epsilon_{\text{motion}} = \tau \cdot l_{\text{motion}} + (1 - \tau) \cdot \epsilon.
\]
The diffusion model $\mathcal{D}_\theta$ takes $l_{\text{VLM}}$, $\epsilon_{\text{motion}}$, and timestep $\tau$ as inputs, aiming to predict the velocity vector field $\epsilon - \epsilon_{\text{motion}}$ for flow matching. To optimize $\mathcal{D}_\theta$, we minimize the following velocity-prediction loss:
\[
\begin{aligned}
\mathcal{L}_{\text{fm}} = \mathbb{E}_{\tau, l_{\text{motion}}, \epsilon} \Bigg[
& \left\| \mathcal{D}_\theta(l_{\text{VLM}}, \epsilon_{\text{motion}}, \tau) \right. \\
& \left. - (\epsilon - \epsilon_{\text{motion}}) \right\|_2^2 \Bigg].
\end{aligned}
\]
This training paradigm ensures that $\mathcal{D}_\theta$ learns to reconstruct motion latents with kinematic information from video semantics, inherently achieving robust cross-modal alignment without separate alignment modules.

\subsection{Policy Training}
\subsubsection{MoE-based Residual Motion Tracker}
We argue that the key to enabling robots to directly observe or infer human motions from videos and perform actions lies in the generalization capability of motion trackers. Specifically, we aim for these trackers to successfully respond to novel prompts while achieving genuine deployment flexibility, which are two critical properties for bridging video understanding and real-world robotic locomotion.

First, we train an oracle teacher policy using the PPO algorithm~\cite{schulman2017proximal} with privileged simulator-state information. To learn a policy $\pi_t$ that generalizes across diverse motion inputs, we first train an initial policy $\pi_0$ on a highly diverse motion dataset $\mathcal{D}_0$~\cite{ji2024exbody2, li2025language}. Given the relative simplicity of egocentric motion datasets $\mathcal D_{\text{first}}$, we restrict the training of $\pi_0$ exclusively to third-person view motion datasets $\mathcal{D}_{\text{third}}$.

Subsequently, we evaluate the tracking accuracy of $\pi_0$ for each motion sequence $s \in \mathcal{D}_{\text{third}}$, with a specific focus on lower-body motion precision. This evaluation employs the error metric $e(s) = \alpha \cdot E_{\text{key}}(s) + \beta \cdot E_{\text{dof}}(s)$, where $E_{\text{key}}(s)$ denotes the mean position error of key lower-body landmarks and $E_{\text{dof}}(s)$ represents the mean tracking error of lower-body joint angles. Motion sequences with $e(s) > 0.6$ are filtered out, and the remaining data $\mathcal{D}$ are used to train a generalizable teacher policy.

The teacher policy $\pi_t$ is trained as a simulation oracle via PPO, utilizing real-world unavailable privileged information: ground-truth root velocity, global joint positions, physical properties (e.g., friction, motor strength), proprioceptive state, and reference motion. Besides, to handle challenging motions, we design the policy to focus on learning dynamic information, specifically a corrective offset $\delta_a$ rather than directly learning kinematic information such as absolute joint targets $p_{\text{target}}$. This offset is added to the joint positions from the reference kinematic trajectory, yielding our final output action $\hat a_t \in \mathbb{R}^{23}$ forming: $\hat a_t = p_{\text{target}} + \delta_a$, 
which is optimized via cumulative rewards to ensure accurate motion tracking and robust behaviors.

Furthermore, we integrate a Mixture of Experts (MoE) module to enhance expressiveness and generalization. The policy includes expert networks and a gating network, which takes the same inputs and computes weights for the experts' outputs to form a weighted sum. The final action is computed as $\hat a_t = \sum_{i=1}^{n} p_i \cdot a_i$, with $p_i$ denoting the gating probability for expert $i$ and $a_i$ is its corresponding output. This design boosts generalization, improves tracking accuracy, and enables precise supervision of the student policy.

\subsubsection{Diffusion-based Student Policy}
Unlike prior work where student policies $\pi_s$ distill knowledge from teachers via explicit reference motion, we regard the student policy as a generation model, which is formulated as a latent-driven generation task~\cite{li2025language}. It takes motion latents generated under video latent guidance as input, alongside observation history, to generate humanoid actions. This design enables the robot to more quickly imitate the motion of the subject in the video, bypassing error-prone steps such as pose estimation and motion retargeting, significantly reducing the time consumption of the entire process.

Following a DAgger-like paradigm, we train the student by rolling it out in simulation, querying the teacher for optimal actions $\hat{a}_t$ at observable states. During training, we inject Gaussian noise $\epsilon_t$ into teacher actions and use our reconstructed latents $l_{\text{v2m}}$, which are from video latents, as guiding latents. The noising process follows a Markov chain:
\begin{equation}
q(x_t | x_{t-1}) = \mathcal{N}(x_t; \sqrt{1 - \alpha_t} \cdot x_{t-1}, \alpha_t \mathbf{I})
\end{equation}
where $\alpha_t \in (0, 1)$ is a sampling hyper-parameter. Denoising at step $t$ is modeled as $x_{t-1} = \epsilon_\theta(x_t, t)$, with $\epsilon_\theta$ as the denoiser. Using an $x_0$-prediction strategy, we supervise via MSE loss: $\mathcal{L} = \| a - \hat{a}_t \|_2^2$, where $a = \frac{x_t - \sqrt{1 - \bar{\alpha}_t} \cdot \epsilon_\theta(x_t, t)}{\sqrt{\bar{\alpha}_t}}$. Converged policies require no privileged knowledge or explicit references, enabling real-world deployment.

\subsubsection{Inference Pipeline}
To ensure fluent, smooth motion, we minimize denoising time by adopting DDIM sampling~\cite{song2020denoising} and an MLP-based diffusion model for action generation. The reverse process is:
\begin{align}
x_{t-1} &= \sqrt{\alpha_{t-1}} \left( \frac{x_t - \sqrt{1 - \alpha_t} \cdot \epsilon_\theta(x_t, t)}{\sqrt{\alpha_t}} \right) \notag \\
&+ \sqrt{1 - \alpha_{t-1}} \cdot \epsilon_\theta(x_t, t)
\end{align}

During inference, our pipeline operates as follows: we first input either an egocentric or third-person video into Qwen3-VL to obtain a video latent representation $l_{\text{vlm}}$. This video latent $l_{\text{vlm}}$ is then fed into our pretrained diffusion model $\mathcal D_{\theta}$, yielding a motion latent $l_{\text{v2m}}$ enriched with kinematic semantics. Finally, we use $l_{\text{v2m}}$ as a conditional signal to guide our diffusion-based student policy, which generates deployable actions directly.

Notably, this pipeline eliminates the need for pose estimation from third-person videos or motion guessing from egocentric videos. By leveraging the video understanding capability of VLMs and the latent comprehension ability of our policy, we successfully achieve video-to-locomotion, enabling the robot to imitate both sparse and dense motion modalities present in the input video.
\section{Experiments}
We evaluate our RoboMirror on both egocentric and third-person videos, aiming to verify its action imitation capabilities for videos of different perspectives respectively. Specifically, for egocentric videos, the model first infers a dense motion latent from the video with sparse motion information, and then performs the imitation task. In the experiments, we train the teacher policy and student policy in the IsaacGym simulation environment, and directly deploy the student policy on the Unitree G1 humanoid robot for real-world testing.

\subsection{Experimental Setups}
\paragraph{Dataset}
We train our model on the Nymeria~\cite{ma2024nymeria} and Motion-X~\cite{lin2023motionx} datasets. Nymeria is a large-scale real-device dataset capturing diverse human daily activities across indoor and outdoor locations, providing paired text-video-motion data—including egocentric videos, full-body motions, and human-annotated motion narrations. This dataset contains approximately 1K videos, each around 15-20 minutes long. Due to its large scale, we select 100 videos, segment each into 5-second clips, and ultimately obtain ~18K video-motion pairs. Both the motions and videos are sampled at 30 FPS.

Motion-X is a large-scale 3D expressive whole-body motion dataset, constructed from massive online videos and eight existing motion datasets, with motions formatted as SMPL-X. We select motion categories paired with videos from it as our third-person dataset. We split the dataset into train and test sets with an 8:2 ratio. 
\paragraph{Metrics}
We adopt two categories of evaluation metrics: motion latent reconstruction and motion tracking. For motion latent reconstruction, we report motion-video retrieval accuracy MV-R@3, MM Dist-V (MM-V), dynamic time warping (DTW), and foot skating to evaluate whether the reconstructed motion latents conform to the semantics of the input videos, where we only evaluate on third-person video datasets. Additionally, we report motion-text retrieval accuracy MT-R@3, MM Dist-T (MM-T), and FID to assess whether the reconstructed motion latents contain sufficient kinematic and semantic information. For motion tracking, evaluated in physics simulators aligning with prior works~\cite{he2024omnih2o}, we use success rate as the core indicator, supplemented by mean per-joint position error ($E_{\text{MPJPE}}$) and mean per-keypoint position error ($E_{\text{MPKPE}}$). Detailed metric definitions are provided in the Appendix.

\paragraph{Implementation Details}
We first input videos into a pretrained VLM for video understanding to obtain video latents, where we use the Qwen3-VL-4B-Instruct model. For the motion latent reconstruction network, we adopt a 16-layer MLP as the backbone and inject conditions via AdaLN~\cite{huang2017arbitrary} to guide the model in reconstructing motion latents from Gaussian noise. For policy training, teacher policy adopts MoE structure of 5 experts, and the student policy employs a diffusion model with a 4-layer MLP as its backbone; this compact network architecture, combined with DDIM sampling, ensures real-time performance during deployment. More details can be found in the Appendix.

\subsection{Evaluation of Motion Latent Reconstruction}
To validate the motion latent reconstruction capability of the diffusion model trained with latents from VLMs as conditions, we decode the reconstructed latents into real motions using our pretrained VAE decoder. The reconstruction performance of motion latents is then evaluated by assessing both the generation quality of the motions and their degree of semantic relevance. Specifically, we evaluate the model's reconstruction performance on the Nymeria and Motion-X datasets, with the results summarized in Table \ref{tab: recon}. 
For video-motion alignment, we further employ MV-R@3, MM-V, DTW~\cite{sakoe2003dynamic}, and foot skating to quantitatively measure the temporal consistency and physical plausibility of reconstructed motions with corresponding videos, which is also presented in Table \ref{tab: recon}.
Here, we introduce a simple baseline, named Vid2Mot, which finetunes the VLM via LoRA~\cite{hu2022lora} while freezing the VAE decoder. This design enables the latents output by the VLM to be directly decoded into motions through the decoder.

\begin{figure*}[t]
\centering
  \includegraphics[width=1.0\textwidth]{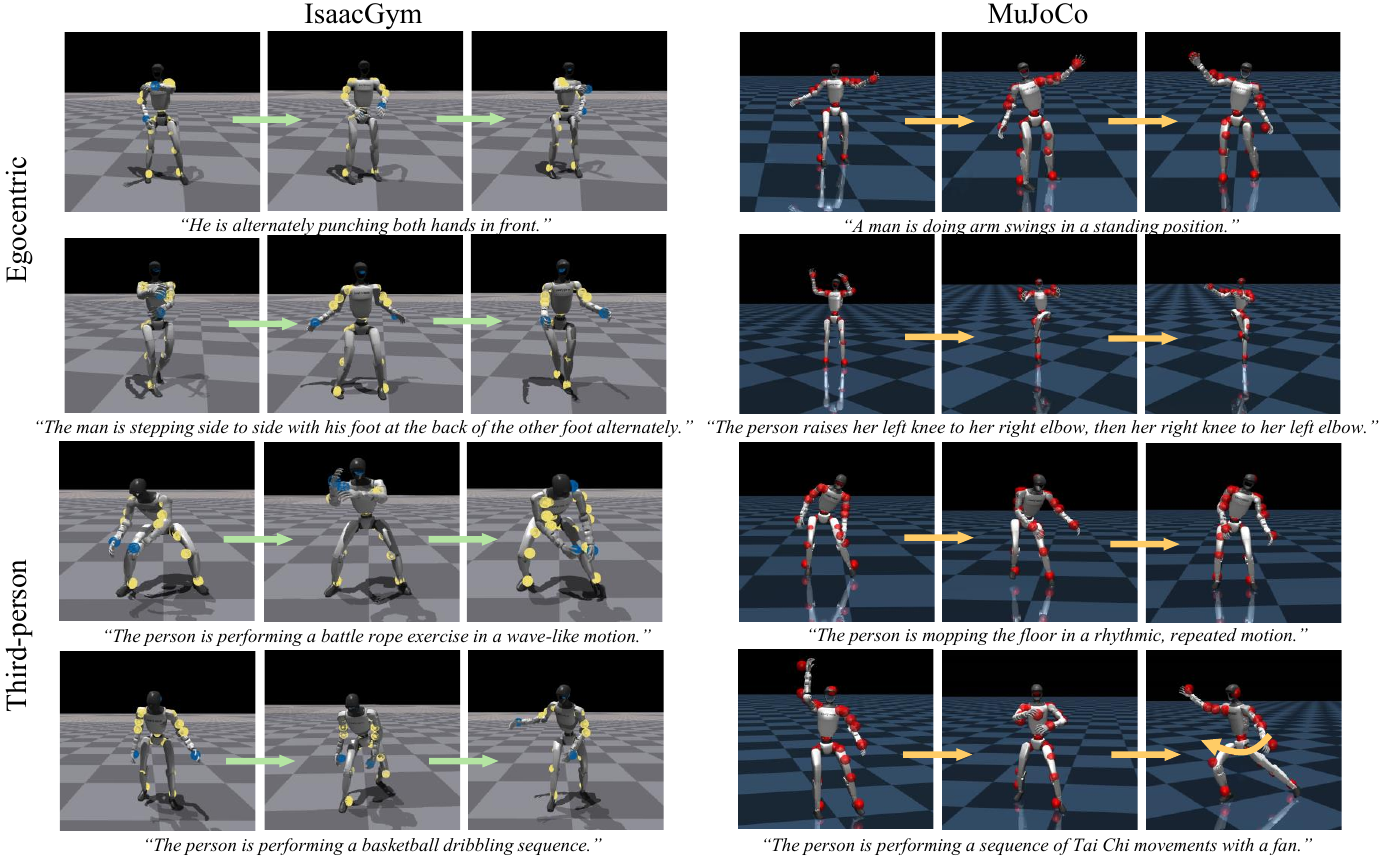}
\caption{Qualitative results in the IsaacGym and MuJoCo. The upper half presents the tracking performance of egocentric videos-to-locomotion, and the lower half presents that of third-person videos-to-locomotion.}
\label{fig:qualitative}
\end{figure*}

\begin{table*}[t] 
\footnotesize   
\setlength{\tabcolsep}{3pt} 
\centering
\resizebox{\linewidth}{!}{
\begin{tabular}{l ccc cccccc}
\toprule
\multirow{2}{*}{Methods} & \multicolumn{3}{c}{Nymeria} & \multicolumn{6}{c}{Motion-X} \\
\cmidrule(lr){2-4} \cmidrule(lr){5-10}
& MT-R@3$\uparrow$ & FID$\downarrow$ & MM-T$\downarrow$ & MT-R@3$\uparrow$ & MV-R@3$\uparrow$ & FID$\downarrow$ & MM-T$\downarrow$ & MM-V$\downarrow$& DTW$\downarrow$ \\
\midrule
Vid2Mot & \cellcolor{color1}{49.34} & \cellcolor{color1}{62.81} & \cellcolor{color1}{25.67} & 69.40 & \cellcolor{color1}{68.8} & 31.27 & 16.22 &\cellcolor{color1}6.65& 0.33 \\
ViMo~\cite{qiu2024vimo} & - & - & - & \cellcolor{color1}{76.92} & 67.8 & \cellcolor{color1}{21.03} & \cellcolor{color1}{8.48} & 7.42 &\cellcolor{color1}{0.31} \\
Ours & \cellcolor{color2}{64.60} & \cellcolor{color2}{42.42} & \cellcolor{color2}{16.97} & \cellcolor{color2}{77.66} & \cellcolor{color2}{72.3} & \cellcolor{color2}{19.53} & \cellcolor{color2}{8.17} & \cellcolor{color2}6.24 &\cellcolor{color2}{0.23} \\
\bottomrule
\end{tabular}
}
\caption{Performance comparison of motion-text and motion-video alignment on the Nymeria and Motion-X datasets.}
\label{tab: recon}
\end{table*}
\begin{table}[t]
\centering
\footnotesize
\setlength{\tabcolsep}{2.8pt}
\renewcommand{\arraystretch}{1.05}

\resizebox{\columnwidth}{!}{
\begin{tabular}{lcccccc}
\toprule
\multirow{2}{*}{Method}
& \multicolumn{3}{c}{IsaacGym}
& \multicolumn{3}{c}{MuJoCo} \\
\cmidrule(lr){2-4} \cmidrule(lr){5-7}
& Succ $\uparrow$
& $E_{\mathrm{mpjpe}}$ $\downarrow$
& $E_{\mathrm{mpkpe}}$ $\downarrow$
& Succ $\uparrow$
& $E_{\mathrm{mpjpe}}$ $\downarrow$
& $E_{\mathrm{mpkpe}}$ $\downarrow$ \\
\midrule

\multicolumn{7}{c}{Nymeria} \\
\midrule
Baseline
& \cellcolor{color1}0.92
& \cellcolor{color1}0.19
& \cellcolor{color1}0.17
& \cellcolor{color1}0.69
& \cellcolor{color1}0.27
& \cellcolor{color1}0.24 \\

Ours
& \cellcolor{color2}0.99
& \cellcolor{color2}0.08
& \cellcolor{color2}0.11
& \cellcolor{color2}0.78
& \cellcolor{color2}0.19
& \cellcolor{color2}0.17 \\

\midrule
\multicolumn{7}{c}{Motion-X} \\
\midrule

Baseline
& \cellcolor{color1}0.91
& \cellcolor{color1}0.23
& \cellcolor{color1}0.20
& \cellcolor{color1}0.61
& \cellcolor{color1}0.36
& \cellcolor{color1}0.33 \\

Ours
& \cellcolor{color2}0.95
& \cellcolor{color2}0.17
& \cellcolor{color2}0.15
& \cellcolor{color2}0.70
& \cellcolor{color2}0.28
& \cellcolor{color2}0.27 \\

\bottomrule
\end{tabular}
}

\caption{Motion tracking performance comparison in simulation on the Nymeria and Motion-X test sets.}
\label{tab:tracking_results}
\end{table}
\begin{table}[t]
\footnotesize   
\setlength{\tabcolsep}{6pt}
\centering
\begin{tabular}{lcccc}
\toprule
Method & Succ $\uparrow$ &R@3 (\%)$\uparrow$& FID $\downarrow$ & MM-Dist$\downarrow$ \\
\midrule
\multicolumn{5}{c}{Nymeria}\\
\midrule
Qwen-VL   & 0.97 &61.1 &46.73 &17.64\\
Qwen2-VL  & \cellcolor{color1}0.98 &61.9 & 44.02 &\cellcolor{color1}17.53\\
Qwen2.5-VL  & 0.97 &\cellcolor{color1}62.6 & \cellcolor{color1}42.74 & 17.80\\
Qwen3-VL  &\cellcolor{color2} 0.99 &\cellcolor{color2}{64.6} & \cellcolor{color2}{42.42}& \cellcolor{color2}{16.97}\\
\midrule
\midrule
\multicolumn{5}{c}{Motion-X}\\
\midrule
Qwen-VL   &0.93& 76.84 & 20.69 & 8.31\\
Qwen2-VL  & 0.94& 77.08 &20.42 & \cellcolor{color1}8.19\\
Qwen2.5-VL  & \cellcolor{color2}0.95&\cellcolor{color1}77.58&\cellcolor{color1}19.77 &8.26\\
Qwen3-VL  &\cellcolor{color2} 0.95&\cellcolor{color2}{77.66}&\cellcolor{color2}{19.53} &\cellcolor{color2}{8.17}\\
\bottomrule
\end{tabular}
\caption{Ablation study on different vision-language models across motion latent reconstruction quality and tracking performance.}
\label{tab:ab1}
\end{table}
\begin{table}[t]
\footnotesize
\centering
\setlength{\tabcolsep}{3.2pt}
\renewcommand{\arraystretch}{1.02}

\begin{tabular}{lcccc}
\toprule
Method
& Succ $\uparrow$
& R@3 (\%) $\uparrow$
& FID $\downarrow$
& MM-Dist $\downarrow$ \\
\midrule

\multicolumn{5}{c}{Nymeria} \\
\midrule
Alignment
& \cellcolor{color1}0.95
& \cellcolor{color1}51.1
& \cellcolor{color1}62.42
& \cellcolor{color1}30.54 \\

Reconstruction
& \cellcolor{color2}0.99
& \cellcolor{color2}64.6
& \cellcolor{color2}42.42
& \cellcolor{color2}16.97 \\

\midrule
\multicolumn{5}{c}{Motion-X} \\
\midrule
Alignment
& \cellcolor{color1}0.92
& \cellcolor{color1}56.41
& \cellcolor{color1}40.19
& \cellcolor{color1}24.37 \\

Reconstruction
& \cellcolor{color2}0.95
& \cellcolor{color2}77.66
& \cellcolor{color2}19.53
& \cellcolor{color2}8.17 \\

\bottomrule
\end{tabular}

\caption{Ablation study on alignment vs.\ reconstruction across motion latent reconstruction quality and tracking performance.}
\label{tab:ab2}
\end{table}
\begin{table}[t]
\footnotesize
\centering
\setlength{\tabcolsep}{2.7pt}
\renewcommand{\arraystretch}{1.02}

\begin{tabular}{lcccc}
\toprule
\multirow{2}{*}{Method}
& \multicolumn{4}{c}{Motion-X} \\
\cmidrule(lr){2-5}
& Succ $\uparrow$
& MPJPE $\downarrow$
& MPKPE $\downarrow$
& Time (s) $\downarrow$ \\
\midrule

Pose-driven
& \cellcolor{color1}0.94
& \cellcolor{color1}0.17
& \cellcolor{color2}0.15
& \cellcolor{color1}9.22 \\

Latent-driven
& \cellcolor{color2}0.95
& \cellcolor{color2}0.16
& \cellcolor{color2}0.15
& \cellcolor{color2}1.84 \\

\bottomrule
\end{tabular}

\caption{Ablation study comparing pose-driven and latent-driven tracking for third-person video-to-locomotion.}
\label{tab:ab3}
\end{table}

\begin{figure}[t]
\centering
  \includegraphics[width=0.8\columnwidth]{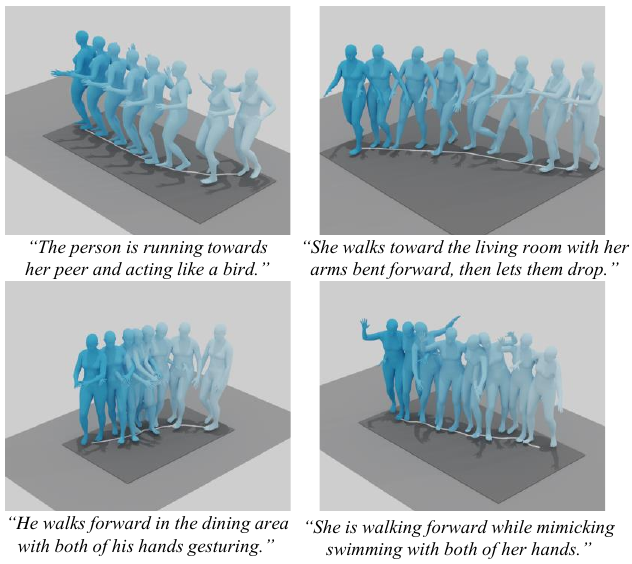}
\caption{Qualitative results of generated motions.}
\label{fig:mogen}
\end{figure}
\vspace{-5mm}
\subsection{Evaluation of Motion Tracking}
To further validate the efficacy of our motion tracking policy, we conduct evaluations on egocentric and third-person videos, measuring the Mean Per Joint Position Error ($E_{\text{mpjpe}}$) and Mean Per Keypoint Position Error ($E_{\text{mpkpe}}$) under physics-based simulations in both IsaacGym and MuJoCo. The pipeline proceeds as follows: first, videos and prompts are fed into the VLM for video understanding and latent representation generation; subsequently, these latents are used as conditions to input into our pretrained diffusion model in Section \ref{vlm} for motion latent reconstruction; finally, the student policy takes this representation and generates actions. As shown in Table \ref{tab:tracking_results}, our method achieves high task success rates on the Nymeria and Motion-X datasets, alongside low joint and keypoint errors---indicating strong alignment between the semantic information of videos and physically executable trajectories. The baseline herein refers to the result of finetuning Qwen3-VL with LoRA, where motion reconstruction is adopted as the optimization objective, and the latents output by Qwen3-VL are directly used to guide the policy in action generation.

\subsection{Qualitative Results}
We conduct a qualitative evaluation of the motion tracking policy across two deployment scenarios: simulation (IsaacGym) and cross-simulator transfer (MuJoCo). Figure \ref{fig:qualitative} illustrates representative tracking sequences, emphasizing the policy's capacity to preserve motion semantics, maintain balance during dynamic transitions, and generalize across distinct physics engines and hardware platforms. 

Additionally, in Figure \ref{fig:mogen}, we also provide visualization results of the generated motion decoded from the motion latents reconstructed from DiT $\mathcal D_{\theta}$ via the pretrained decoder.

\subsection{Ablation Studies}

To systematically validate the effectiveness of our method, we conduct a series of ablation studies in this section. These experiments cover three key aspects: 1) different video understanding models, 2) diverse approaches for converting video latents to motion latents, and 3) the advantages of our method over pose estimation-based student policy for third-person video-driven locomotion.

\paragraph{Different Vision-Language Models} To verify the impact of different vision-language models on the final results, we employ four distinct models: Qwen-VL, Qwen-2-VL, Qwen2.5-VL, and Qwen3-VL. We evaluate their performance in terms of motion reconstruction quality and their influence on policy tracking performance, with the results summarized in Table \ref{tab:ab1}. As indicated in the table, Qwen3-VL generates superior latent representations that are more suitable for motion latent reconstruction, thereby enabling more effective control of humanoid robots.

\paragraph{Alignment Vs Reconstruction} 
We argue that reconstruction outperforms alignment. Since the latents output by VLM lack kinematic information, directly using them to guide action generation results in unstable motions with ambiguous task relevance. Thus, it is necessary to convert $l_{\text{vlm}}$ into $l_{\text{motion}}$ embedded with kinematic cues. One approach is to align $l_{\text{vlm}}$ with $l_{\text{motion}}$ to impart kinematic information, which then guides action generation. However, we hypothesize that reconstructing target motion latents $l_{\text{motion}}$ from VLM latents $l_{\text{vlm}}$ can more effectively bridge the domain gap between visual semantics and motion dynamics.

To validate this, we conduct ablation experiments on alignment and reconstruction. For the alignment baseline, we train a 4-layer transformer adapter with InfoNCE loss, which pulls positive sample pairs closer while pushing negative pairs apart. As shown in Table \ref{tab:ab2}, reconstructing $l_{\text{motion}}$ from $l_{\text{vlm}}$ achieves significantly superior performance.

\paragraph{Latent-driven Vs Pose-driven}
For evaluating our method's ability to imitate actions from third-person videos, we assess the advantages of RoboMirror over conventional pose estimation-based methods. Herein, pose estimation-based methods refer to approaches that first estimate poses from video test sets, then retarget the poses to the humanoid, and finally feed the retargeted reference motion as input to the student policy for action generation. As shown in Table \ref{tab:ab3}, RoboMirror achieves superior tracking performance with lower overall pipeline latency. This is attributed to the fact that both pose estimation and retargeting incur non-negligible time costs and introduce cumulative errors, whereas our framework avoids such inefficiencies.

\section{Conclusion}

We present RoboMirror, a video-to-locomotion framework rooted in "understand before you imitate." Leveraging VLMs, it extracts semantic motion intents from videos and reconstructs them into kinematically grounded latents, enabling humanoids to generate physically plausible, semantically aligned actions without pose estimation or retargeting. Extensive experiments validate its superiority in task success, latency, and cross-domain generalization. RoboMirror bridges visual understanding and humanoid locomotion, laying groundwork for understanding-driven control.

\section{Acknowledgement}
This work was supported by the National Natural Science Foundation of China (62476011) and Beijing Natural Science Foundation (L252060).

\bibliographystyle{assets/plainnat}
\bibliography{paper}

@String(ICLR  = {Int. Conf. Learn. Represent.})

@String(TOG   = {ACM Trans. Graph.})

@String(ICLR  = {ICLR})

@String(TOG   = {ACM TOG})

@article{Qwen2.5-VL,
  title={Qwen2.5-VL Technical Report},
  author={Bai, Shuai and Chen, Keqin and Liu, Xuejing and Wang, Jialin and Ge, Wenbin and Song, Sibo and Dang, Kai and Wang, Peng and Wang, Shijie and Tang, Jun and Zhong, Humen and Zhu, Yuanzhi and Yang, Mingkun and Li, Zhaohai and Wan, Jianqiang and Wang, Pengfei and Ding, Wei and Fu, Zheren and Xu, Yiheng and Ye, Jiabo and Zhang, Xi and Xie, Tianbao and Cheng, Zesen and Zhang, Hang and Yang, Zhibo and Xu, Haiyang and Lin, Junyang},
  journal={arXiv preprint arXiv:2502.13923},
  year={2025}
}

@article{kingma2013auto,
  title={Auto-encoding variational bayes},
  author={Kingma, Diederik P and Welling, Max},
  journal={arXiv preprint arXiv:1312.6114},
  year={2013}
}

@article{bjorck2025gr00t,
  title={Gr00t n1: An open foundation model for generalist humanoid robots},
  author={Bjorck, Johan and Casta{\~n}eda, Fernando and Cherniadev, Nikita and Da, Xingye and Ding, Runyu and Fan, Linxi and Fang, Yu and Fox, Dieter and Hu, Fengyuan and Huang, Spencer and others},
  journal={arXiv preprint arXiv:2503.14734},
  year={2025}
}

@article{schulman2017proximal,
  title={Proximal policy optimization algorithms},
  author={Schulman, John and Wolski, Filip and Dhariwal, Prafulla and Radford, Alec and Klimov, Oleg},
  journal={arXiv preprint arXiv:1707.06347},
  year={2017}
}

@article{li2025language,
  title={From Language to Locomotion: Retargeting-free Humanoid Control via Motion Latent Guidance},
  author={Li, Zhe and Chi, Cheng and Wei, Yangyang and Zhu, Boan and Peng, Yibo and Huang, Tao and Wang, Pengwei and Wang, Zhongyuan and Zhang, Shanghang and Xu, Chang},
  journal={arXiv preprint arXiv:2510.14952},
  year={2025}
}

@article{weng2025hdmi,
  title={HDMI: Learning Interactive Humanoid Whole-Body Control from Human Videos},
  author={Weng, Haoyang and Li, Yitang and Sobanbabu, Nikhil and Wang, Zihan and Luo, Zhengyi and He, Tairan and Ramanan, Deva and Shi, Guanya},
  journal={arXiv preprint arXiv:2509.16757},
  year={2025}
}

@article{song2020denoising,
  title={Denoising diffusion implicit models},
  author={Song, Jiaming and Meng, Chenlin and Ermon, Stefano},
  journal={arXiv preprint arXiv:2010.02502},
  year={2020}
}

@inproceedings{ma2024nymeria,
  title={Nymeria: A massive collection of multimodal egocentric daily motion in the wild},
  author={Ma, Lingni and Ye, Yuting and Hong, Fangzhou and Guzov, Vladimir and Jiang, Yifeng and Postyeni, Rowan and Pesqueira, Luis and Gamino, Alexander and Baiyya, Vijay and Kim, Hyo Jin and others},
  booktitle={European Conference on Computer Vision},
  pages={445--465},
  year={2024},
  organization={Springer}
}

@article{lin2023motionx,
  title={Motion-X: A Large-scale 3D Expressive Whole-body Human Motion Dataset},
  author={Lin, Jing and Zeng, Ailing and Lu, Shunlin and Cai, Yuanhao and Zhang, Ruimao and Wang, Haoqian and Zhang, Lei},
  journal={Advances in Neural Information Processing Systems},
  year={2023}
}

@inproceedings{huang2017arbitrary,
  title={Arbitrary style transfer in real-time with adaptive instance normalization},
  author={Huang, Xun and Belongie, Serge},
  booktitle={Proceedings of the IEEE international conference on computer vision},
  pages={1501--1510},
  year={2017}
}

@article{qiu2024vimo,
  title={Vimo: Generating motions from casual videos},
  author={Qiu, Liangdong and Yu, Chengxing and Li, Yanran and Wang, Zhao and Huang, Haibin and Ma, Chongyang and Zhang, Di and Wan, Pengfei and Han, Xiaoguang},
  journal={arXiv preprint arXiv:2408.06614},
  year={2024}
}

@article{li2024lamp,
  title={Lamp: Language-motion pretraining for motion generation, retrieval, and captioning},
  author={Li, Zhe and Yuan, Weihao and He, Yisheng and Qiu, Lingteng and Zhu, Shenhao and Gu, Xiaodong and Shen, Weichao and Dong, Yuan and Dong, Zilong and Yang, Laurence T},
  journal={arXiv preprint arXiv:2410.07093},
  year={2024}
}

@article{li2024mulsmo,
  title={Mulsmo: Multimodal stylized motion generation by bidirectional control flow},
  author={Li, Zhe and He, Yisheng and Zhong, Lei and Shen, Weichao and Zuo, Qi and Qiu, Lingteng and Dong, Zilong and Yang, Laurence Tianruo and Yuan, Weihao},
  journal={arXiv preprint arXiv:2412.09901},
  year={2024}
}

@inproceedings{guo2024momask,
  title={Momask: Generative masked modeling of 3d human motions},
  author={Guo, Chuan and Mu, Yuxuan and Javed, Muhammad Gohar and Wang, Sen and Cheng, Li},
  booktitle={Proceedings of the IEEE/CVF Conference on Computer Vision and Pattern Recognition},
  pages={1900--1910},
  year={2024}
}

@article{he2025asap,
  title={Asap: Aligning simulation and real-world physics for learning agile humanoid whole-body skills},
  author={He, Tairan and Gao, Jiawei and Xiao, Wenli and Zhang, Yuanhang and Wang, Zi and Wang, Jiashun and Luo, Zhengyi and He, Guanqi and Sobanbab, Nikhil and Pan, Chaoyi and others},
  journal={arXiv preprint arXiv:2502.01143},
  year={2025}
}

@article{xie2025kungfubot,
  title={KungfuBot: Physics-Based Humanoid Whole-Body Control for Learning Highly-Dynamic Skills},
  author={Xie, Weiji and Han, Jinrui and Zheng, Jiakun and Li, Huanyu and Liu, Xinzhe and Shi, Jiyuan and Zhang, Weinan and Bai, Chenjia and Li, Xuelong},
  journal={arXiv preprint arXiv:2506.12851},
  year={2025}
}

@inproceedings{serifi2024robot,
  title={Robot motion diffusion model: Motion generation for robotic characters},
  author={Serifi, Agon and Grandia, Ruben and Knoop, Espen and Gross, Markus and B{\"a}cher, Moritz},
  booktitle={SIGGRAPH asia 2024 conference papers},
  pages={1--9},
  year={2024}
}

@article{Qwen2-VL,
  title={Qwen2-VL: Enhancing Vision-Language Model's Perception of the World at Any Resolution},
  author={Wang, Peng and Bai, Shuai and Tan, Sinan and Wang, Shijie and Fan, Zhihao and Bai, Jinze and Chen, Keqin and Liu, Xuejing and Wang, Jialin and Ge, Wenbin and Fan, Yang and Dang, Kai and Du, Mengfei and Ren, Xuancheng and Men, Rui and Liu, Dayiheng and Zhou, Chang and Zhou, Jingren and Lin, Junyang},
  journal={arXiv preprint arXiv:2409.12191},
  year={2024}
}

@article{Qwen-VL,
  title={Qwen-VL: A Versatile Vision-Language Model for Understanding, Localization, Text Reading, and Beyond},
  author={Bai, Jinze and Bai, Shuai and Yang, Shusheng and Wang, Shijie and Tan, Sinan and Wang, Peng and Lin, Junyang and Zhou, Chang and Zhou, Jingren},
  journal={arXiv preprint arXiv:2308.12966},
  year={2023}
}

@article{hu2022lora,
  title={Lora: Low-rank adaptation of large language models.},
  author={Hu, Edward J and Shen, Yelong and Wallis, Phillip and Allen-Zhu, Zeyuan and Li, Yuanzhi and Wang, Shean and Wang, Lu and Chen, Weizhu and others},
  journal={ICLR},
  volume={1},
  number={2},
  pages={3},
  year={2022}
}

@article{he2024omnih2o,
  title={Omnih2o: Universal and dexterous human-to-humanoid whole-body teleoperation and learning},
  author={He, Tairan and Luo, Zhengyi and He, Xialin and Xiao, Wenli and Zhang, Chong and Zhang, Weinan and Kitani, Kris and Liu, Changliu and Shi, Guanya},
  journal={arXiv preprint arXiv:2406.08858},
  year={2024}
}

@article{ji2024exbody2,
  title={Exbody2: Advanced expressive humanoid whole-body control},
  author={Ji, Mazeyu and Peng, Xuanbin and Liu, Fangchen and Li, Jialong and Yang, Ge and Cheng, Xuxin and Wang, Xiaolong},
  journal={arXiv preprint arXiv:2412.13196},
  year={2024}
}

@article{chen2025gmt,
  title={GMT: General Motion Tracking for Humanoid Whole-Body Control},
  author={Chen, Zixuan and Ji, Mazeyu and Cheng, Xuxin and Peng, Xuanbin and Peng, Xue Bin and Wang, Xiaolong},
  journal={arXiv preprint arXiv:2506.14770},
  year={2025}
}

@article{yue2025rl,
  title={RL from Physical Feedback: Aligning Large Motion Models with Humanoid Control},
  author={Yue, Junpeng and Wang, Zepeng and Wang, Yuxuan and Zeng, Weishuai and Wang, Jiangxing and Xu, Xinrun and Zhang, Yu and Zheng, Sipeng and Ding, Ziluo and Lu, Zongqing},
  journal={arXiv preprint arXiv:2506.12769},
  year={2025}
}

@article{shao2025langwbc,
  title={LangWBC: Language-directed Humanoid Whole-Body Control via End-to-end Learning},
  author={Shao, Yiyang and Huang, Xiaoyu and Zhang, Bike and Liao, Qiayuan and Gao, Yuman and Chi, Yufeng and Li, Zhongyu and Shao, Sophia and Sreenath, Koushil},
  journal={arXiv preprint arXiv:2504.21738},
  year={2025}
}

@article{han2025kungfubot2,
  title={KungfuBot2: Learning Versatile Motion Skills for Humanoid Whole-Body Control},
  author={Han, Jinrui and Xie, Weiji and Zheng, Jiakun and Shi, Jiyuan and Zhang, Weinan and Xiao, Ting and Bai, Chenjia},
  journal={arXiv preprint arXiv:2509.16638},
  year={2025}
}

@article{li2025omnimotion,
  title={OmniMotion: Multimodal Motion Generation with Continuous Masked Autoregression},
  author={Li, Zhe and Yuan, Weihao and Shen, Weichao and Zhu, Siyu and Dong, Zilong and Xu, Chang},
  journal={arXiv preprint arXiv:2510.14954},
  year={2025}
}

@article{peng2018deepmimic,
  title={Deepmimic: Example-guided deep reinforcement learning of physics-based character skills},
  author={Peng, Xue Bin and Abbeel, Pieter and Levine, Sergey and Van de Panne, Michiel},
  journal={ACM Transactions On Graphics (TOG)},
  volume={37},
  number={4},
  pages={1--14},
  year={2018},
  publisher={ACM New York, NY, USA}
}

@article{zhang2025hub,
  title={HuB: Learning Extreme Humanoid Balance},
  author={Zhang, Tong and Zheng, Boyuan and Nai, Ruiqian and Hu, Yingdong and Wang, Yen-Jen and Chen, Geng and Lin, Fanqi and Li, Jiongye and Hong, Chuye and Sreenath, Koushil and others},
  journal={arXiv preprint arXiv:2505.07294},
  year={2025}
}

@article{ze2025twist,
  title={Twist: Teleoperated whole-body imitation system},
  author={Ze, Yanjie and Chen, Zixuan and Ara{\'u}jo, Joao Pedro and Cao, Zi-ang and Peng, Xue Bin and Wu, Jiajun and Liu, C Karen},
  journal={arXiv preprint arXiv:2505.02833},
  year={2025}
}

@article{li2025clone,
  title={CLONE: Closed-Loop Whole-Body Humanoid Teleoperation for Long-Horizon Tasks},
  author={Li, Yixuan and Lin, Yutang and Cui, Jieming and Liu, Tengyu and Liang, Wei and Zhu, Yixin and Huang, Siyuan},
  journal={arXiv preprint arXiv:2506.08931},
  year={2025}
}

@article{wang2025experts,
  title={From experts to a generalist: Toward general whole-body control for humanoid robots},
  author={Wang, Yuxuan and Yang, Ming and Ding, Ziluo and Zhang, Yu and Zeng, Weishuai and Xu, Xinrun and Jiang, Haobin and Lu, Zongqing},
  journal={arXiv preprint arXiv:2506.12779},
  year={2025}
}

@article{yin2025unitracker,
  title={Unitracker: Learning universal whole-body motion tracker for humanoid robots},
  author={Yin, Kangning and Zeng, Weishuai and Fan, Ke and Dai, Minyue and Wang, Zirui and Zhang, Qiang and Tian, Zheng and Wang, Jingbo and Pang, Jiangmiao and Zhang, Weinan},
  journal={arXiv preprint arXiv:2507.07356},
  year={2025}
}

@article{liao2025beyondmimic,
  title={Beyondmimic: From motion tracking to versatile humanoid control via guided diffusion},
  author={Liao, Qiayuan and Truong, Takara E and Huang, Xiaoyu and Tevet, Guy and Sreenath, Koushil and Liu, C Karen},
  journal={arXiv preprint arXiv:2508.08241},
  year={2025}
}

@article{geyer2003positive,
  title={Positive force feedback in bouncing gaits?},
  author={Geyer, Hartmut and Seyfarth, Andre and Blickhan, Reinhard},
  journal={Proceedings of the Royal Society of London. Series B: Biological Sciences},
  volume={270},
  number={1529},
  pages={2173--2183},
  year={2003},
  publisher={The Royal Society}
}

@article{sreenath2011compliant,
  title={A compliant hybrid zero dynamics controller for stable, efficient and fast bipedal walking on MABEL},
  author={Sreenath, Koushil and Park, Hae-Won and Poulakakis, Ioannis and Grizzle, Jessy W},
  journal={The International Journal of Robotics Research},
  volume={30},
  number={9},
  pages={1170--1193},
  year={2011},
  publisher={SAGE Publications Sage UK: London, England}
}

@article{wang2025beamdojo,
  title={Beamdojo: Learning agile humanoid locomotion on sparse footholds},
  author={Wang, Huayi and Wang, Zirui and Ren, Junli and Ben, Qingwei and Huang, Tao and Zhang, Weinan and Pang, Jiangmiao},
  journal={arXiv preprint arXiv:2502.10363},
  year={2025}
}

@article{li2023robust,
  title={Robust and versatile bipedal jumping control through reinforcement learning},
  author={Li, Zhongyu and Peng, Xue Bin and Abbeel, Pieter and Levine, Sergey and Berseth, Glen and Sreenath, Koushil},
  journal={arXiv preprint arXiv:2302.09450},
  year={2023}
}

@article{huang2025learning,
  title={Learning humanoid standing-up control across diverse postures},
  author={Huang, Tao and Ren, Junli and Wang, Huayi and Wang, Zirui and Ben, Qingwei and Wen, Muning and Chen, Xiao and Li, Jianan and Pang, Jiangmiao},
  journal={arXiv preprint arXiv:2502.08378},
  year={2025}
}

@article{he2025learning,
  title={Learning getting-up policies for real-world humanoid robots},
  author={He, Xialin and Dong, Runpei and Chen, Zixuan and Gupta, Saurabh},
  journal={arXiv preprint arXiv:2502.12152},
  year={2025}
}

@inproceedings{li2025hold,
  title={Hold My Beer: Learning Gentle Humanoid Locomotion and End-Effector Stabilization Control},
  author={Li, Yitang and Zhang, Yuanhang and Xiao, Wenli and Pan, Chaoyi and Weng, Haoyang and He, Guanqi and He, Tairan and Shi, Guanya},
  booktitle={RSS 2025 Workshop on Whole-body Control and Bimanual Manipulation: Applications in Humanoids and Beyond}
}

@article{zhang2025falcon,
  title={FALCON: Learning Force-Adaptive Humanoid Loco-Manipulation},
  author={Zhang, Yuanhang and Yuan, Yifu and Gurunath, Prajwal and He, Tairan and Omidshafiei, Shayegan and Agha-mohammadi, Ali-akbar and Vazquez-Chanlatte, Marcell and Pedersen, Liam and Shi, Guanya},
  journal={arXiv preprint arXiv:2505.06776},
  year={2025}
}

@article{su2025hitter,
  title={Hitter: A humanoid table tennis robot via hierarchical planning and learning},
  author={Su, Zhi and Zhang, Bike and Rahmanian, Nima and Gao, Yuman and Liao, Qiayuan and Regan, Caitlin and Sreenath, Koushil and Sastry, S Shankar},
  journal={arXiv preprint arXiv:2508.21043},
  year={2025}
}

@article{peng2021amp,
  title={Amp: Adversarial motion priors for stylized physics-based character control},
  author={Peng, Xue Bin and Ma, Ze and Abbeel, Pieter and Levine, Sergey and Kanazawa, Angjoo},
  journal={ACM Transactions on Graphics (ToG)},
  volume={40},
  number={4},
  pages={1--20},
  year={2021},
  publisher={ACM New York, NY, USA}
}

@misc{xue2025leverbhumanoidwholebodycontrol,
      title={LeVERB: Humanoid Whole-Body Control with Latent Vision-Language Instruction}, 
      author={Haoru Xue and Xiaoyu Huang and Dantong Niu and Qiayuan Liao and Thomas Kragerud and Jan Tommy Gravdahl and Xue Bin Peng and Guanya Shi and Trevor Darrell and Koushil Sreenath and Shankar Sastry},
      year={2025},
      eprint={2506.13751},
      archivePrefix={arXiv},
      primaryClass={cs.RO},
      url={https://arxiv.org/abs/2506.13751}, 
}

@article{sakoe2003dynamic,
  title={Dynamic programming algorithm optimization for spoken word recognition},
  author={Sakoe, Hiroaki and Chiba, Seibi},
  journal={IEEE transactions on acoustics, speech, and signal processing},
  volume={26},
  number={1},
  pages={43--49},
  year={2003},
  publisher={IEEE}
}

@inproceedings{li2026you,
  title={Do you have freestyle? expressive humanoid locomotion via audio control},
  author={Li, Zhe and Chi, Cheng and Wei, Yangyang and Zhu, Boan and Huang, Tao and Sun, Zhenguo and Peng, Yibo and Wang, Pengwei and Wang, Zhongyuan and Liu, Fangzhou and others},
  booktitle={Proceedings of the IEEE/CVF Conference on Computer Vision and Pattern Recognition},
  pages={956--965},
  year={2026}
}

@article{yuan2026roboforge,
  title={RoboForge: Physically optimized text-guided whole-body locomotion for humanoids},
  author={Yuan, Xichen and Li, Zhe and Lyu, Bofan and Zuo, Kuangji and Lu, Yanshuo and Li, Gen and Yang, Jianfei},
  journal={arXiv preprint arXiv:2603.17927},
  year={2026}
}

@inproceedings{li2023enhancing,
  title={Enhancing sentence representation with visually-supervised multimodal pre-training},
  author={Li, Zhe and Yang, Laurence T and Nie, Xin and Ren, BoCheng and Deng, Xianjun},
  booktitle={Proceedings of the 31st ACM International Conference on Multimedia},
  pages={5686--5695},
  year={2023}
}

@article{tan2026robobrain,
    title={RoboBrain 2.5: Depth in Sight, Time in Mind},
    author={Tan, Huajie and Zhou, Enshen and Li, Zhiyu and Xu, Yijie and Ji, Yuheng and Chen, Xiansheng and Chi, Cheng and Wang, Pengwei and Jia, Huizhu and Ao, Yulong and others},
    journal={arXiv preprint arXiv:2601.14352},
    year={2026}
}

@article{zhou2025roborefer,
    title={RoboRefer: Towards Spatial Referring with Reasoning in Vision-Language Models for Robotics},
    author={Zhou, Enshen and An, Jingkun and Chi, Cheng and Han, Yi and Rong, Shanyu and Zhang, Chi and Wang, Pengwei and Wang, Zhongyuan and Huang, Tiejun and Sheng, Lu and others},
    journal={arXiv preprint arXiv:2506.04308},
    year={2025}
}

@article{zhou2025robotracer,
    title={RoboTracer: Mastering Spatial Trace with Reasoning in Vision-Language Models for Robotics},
    author={Zhou, Enshen and Chi, Cheng and Li, Yibo and An, Jingkun and Zhang, Jiayuan and Rong, Shanyu and Han, Yi and Ji, Yuheng and Liu, Mengzhen and Wang, Pengwei and others},
    journal={arXiv preprint arXiv:2512.13660},
    year={2025}
}

@article{liu2026sapave,
    title={SaPaVe: Towards Active Perception and Manipulation in Vision-Language-Action Models for Robotics},
    author={Liu, Mengzhen and Zhou, Enshen and Chi, Cheng and Han, Yi and Rong, Shanyu and Chen, Liming and Wang, Pengwei and Wang, Zhongyuan and Zhang, Shanghang},
    journal={arXiv preprint arXiv:2603.12193},
    year={2026}
}
\end{document}